# DeepFusionNet: Autoencoder-Based Low-Light Image Enhancement and Super-Resolution

Halil Hüseyin Çalışkan[*1], Talha Koruk[1]

[1]Department of Computer Engineering, Faculty of Engineering and Natural Sciences, Bursa Technical University,Turkiye

[1*]```caliskanhalil815@gmail.com```
[1]```talha.koruk@btu.edu.tr```

*Abstract*— Computer vision and image processing applications suffer from dark and low-light images, particularly during real-time image transmission. Currently, low light and dark images are converted to bright and colored forms using autoencoders; however, these methods often achieve low SSIM and PSNR scores and require high computational power due to their large number of parameters. To address these challenges, the DeepFusionNet architecture has been developed. According to the results obtained with the LOL-v1 dataset, DeepFusionNet achieved an SSIM of 92.8% and a PSNR score of 26.30, while containing only approximately 2.5 million parameters. On the other hand, conversion of blurry and low-resolution images into high-resolution and blur-free images has gained importance in image processing applications. Unlike GAN-based super-resolution methods, an autoencoder-based super-resolution model has been developed that contains approximately 100 thousand parameters and uses the DeepFusionNet architecture. According to the results of the tests, the DeepFusionNet based super-resolution method achieved a PSNR of 25.30 and a SSIM score of 80.7 percent according to the validation set.

## I. Introduction

Computer vision applications such as object detection, object tracking and object segmentation, and image processing applications such as image transfer and transmission are performed in environments where lighting conditions and image resolutions are suitable and convenient. However, these computer vision and image processing activities are also required to be performed at night or in low-light environments with low computational power for security and surveillance [1]. Additionally, increasing the resolution of images by removing blur and noise has become important in areas such as surveillance.

Machine learning models with linear solutions based on matrix multiplication, scalar multiplication, Euclidean-assisted distance measurement, or Gaussian probability density are inadequate for such complex nonlinear processing [2]. Therefore, artificial neural networks supplemented with activation functions are used in deep learning to enable the model to learn linear relationships. However, artificial neural networks struggle to extract features from images with three channels and large dimensions. To overcome these problems encountered in feature extraction in images, CNN layers that perform convolution operations on the image have been introduced [3]. Currently, the functions and structures of CNN layers have changed, and new structures have emerged such as Autoencoder, Generators and Discriminators.

Autoencoders basically consist of three parts: encoder, bottleneck and decoder [4]. The encoder compresses the input into a lower-dimensional representation in the bottleneck, known as the latent space, which best summarizes and represents the essential features of the data. The information from the latent space is decompressed by the decoder, producing the image output. One of the models based on this autoencoder architecture is the U-Net architecture. Unlike the skip connection method used in ResNet18, which involves aggregation, the U-Net architecture allows the information from the encoder to be directly transmitted to the decoder and combined on a channel-by-channel basis [5]. The DeepFusionNet model is based on many architectures such as U-Net, Resnet18, and DenseNet in terms of feature extraction and combination.

## II. Methods

The DeepFusionNet architecture consists of three main parts: an encoder, a bottleneck, and a decoder. The input to the encoder passes through CNN layers with two different kernel sizes (3×3, 5×5). The principle of passing input through CNNs with different kernel sizes is used in Inception Blocks [6]. This allows the model to extract information at both local and spatial levels. Many activation functions were examined when designing the DeepFusionNet architecture [7]. If the input to the ReLU activation function is positive, it returns itself, while negative inputs return 0. When taking the derivative of ReLU, positive inputs return 1, while negative inputs return 0. Therefore, it is easy to calculate. Unlike Tanh and Silu, it mitigates the gradient problem. However, in ReLU, there is no weight update for gradients with a value of 0, resulting in the Dying Relu problem, also known as the dead neuron, where the model stops learning. The LeakyReLU function was developed to address these challenges encountered in ReLU. In LeakyReLU, negative values are multiplied by a small gradient, resulting in less gradient loss than in ReLU. However, the alpha gradient angle left for negative values is manually adjusted, and this alpha is not updated during training. In the Tanh function, when the inputs are very small or very large, the derivative approaches 0, and the weights are not updated. Furthermore, the computation is slower than ReLU and LeakyReLU because it uses an exponential process [13]. The ReLU6 function, on the other hand, limits positive numbers greater than 6 in ReLU to 6, but because BatchNorm2d is used in the DeepFusionNet architecture, the inputs are normalized to have a mean of 0 and a variance of 1, thus limiting the functionality of ReLU6. In the Silu function, the output is the multiplication of the input itself and its sigmoidal state. This results in smoother transitions than in ReLU. However, the gradient loss observed in Tanh is also observed in Silu; the derivative of very small negative values approaches 0, causing the weights not to be updated. The gradient loss in the PReLU activation function is lower than in other activation functions because positive inputs are passed through unchanged, while negative inputs are scaled by a learnable parameter, alpha, which is continuously updated during each epoch. During gradient computation, the derivative is 1 for positive inputs and alpha for negative inputs. This mechanism helps prevent gradient explosion, a problem that arises when gradients grow excessively large. These characteristics make PReLU particularly suitable as the primary activation function in the DeepFusionNet architecture. Outputs from CNNs with two different kernel sizes and PReLU pass through the Depthwise Convolution (DWCNN) layer, reducing the number of parameters compared to standard CNNs [8]. The outputs from the architecture's 3x3 CNN and the DWCNN are combined using the channel-based concatenation process used in the U-Net architecture. The outputs from the architecture's 5x5 CNN and the DWCNN are passed through the PReLU function again, and the output is combined channel-wise with the outputs from the 5x5 CNN and PReLU. The outputs from both branches are combined channel-wise. The outputs of the two concat operations are then combined channel-wise. As a result of all these operations, both local and spatial feature extraction are achieved. The GhostConv layer is formed by combining CNN and depthwise layers, reducing the number of parameters [9]. The final concat operation learns the relationship between the passed and unpassed values from the activation function on a channel-by-channel basis. Figure 1 shows an example of the architecture up to this point.

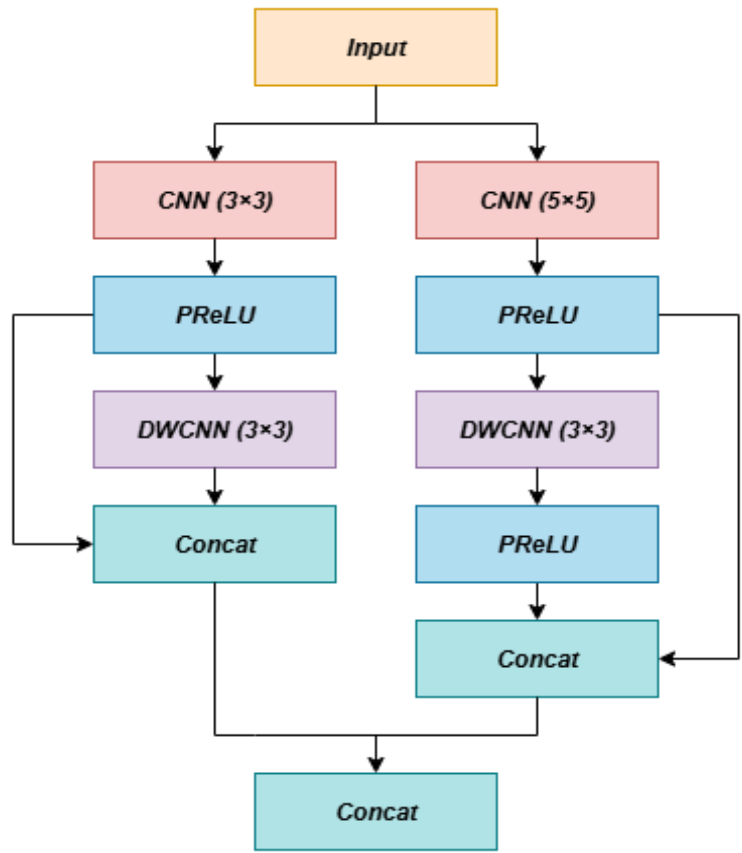

Figure 1. A representation of head of the DeepFusionNet

The resulting output is then passed through the CNN (3x3), PReLU, and BatchNorm2d, and is ready for the CBAM layer. The ChannelAttention layer within the CBAM layer determines which channels are more important as a result of the Concat and subsequent convolution operations, and assigns higher weights to the more important layers [10]. The outputs from ChannelAttention are passed through Spatial Attention, and instead of channel-by-channel analysis, spatial regions are learned to be more important. There is no dimension change in the CBAM layer; only the features are reweighted. Then, dimensionality reduction is performed with MaxPool2d, and deeper features are learned using the sequential CNN and activation function logic used in the ResNet18 architecture. These features are normalized with BathNorm2d and made available in the CBAM layer. This process continues until the bottleneck of the DeepFusionNet architecture. The entire encoder architecture of DeepFusionNet, from the head to the bottleneck, is shown in Figure 2.

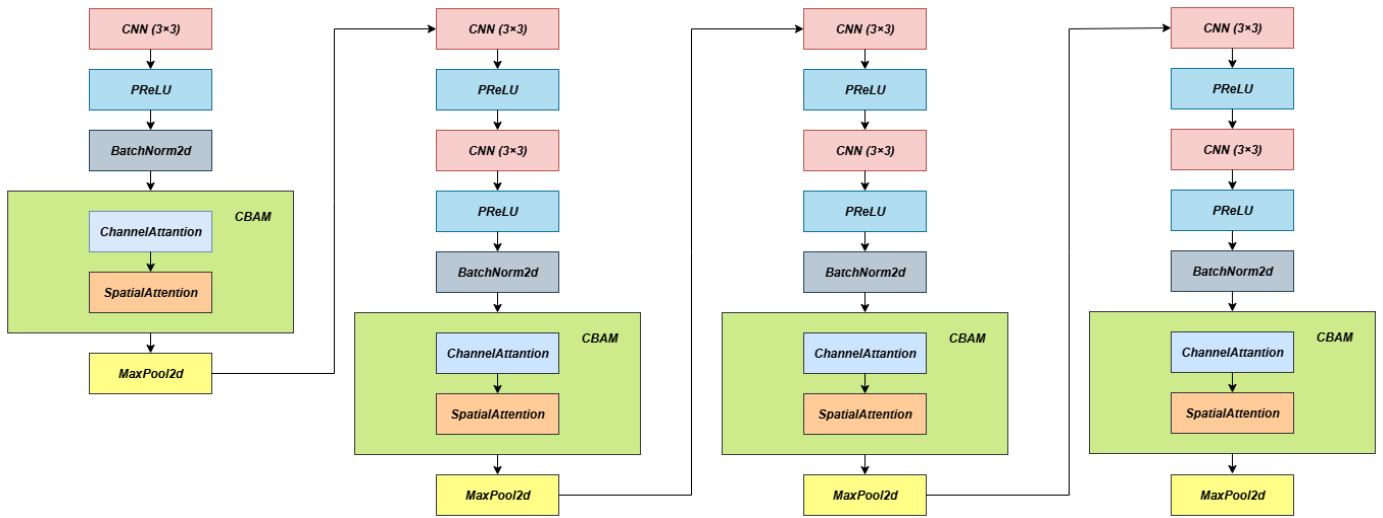

Figure 2. Demonstration of the entire encoder architecture of DeepFusionNet, from the head to the bottleneck

In the bottleneck, the number of channels is reduced using a Pointwise CNN (1×1), and feature extraction is performed sequentially from the CNNs (3x3) [11]. All CNN outputs are combined on a channel-by-channel basis using an approach similar to the DenseNet architecture. This creates a latent space containing the information necessary for the model to decode. The main reason for not using the inverted bottleneck used in Mobilenet-V2 is to ensure sufficient feature extraction for the latent space rather than performance [15]. Then, the CBAM layers are reused, so the weights vary according to their importance not only for the encoder but also within the bottleneck. A schematic representation of the Bottleneck is shown in Figure 3.

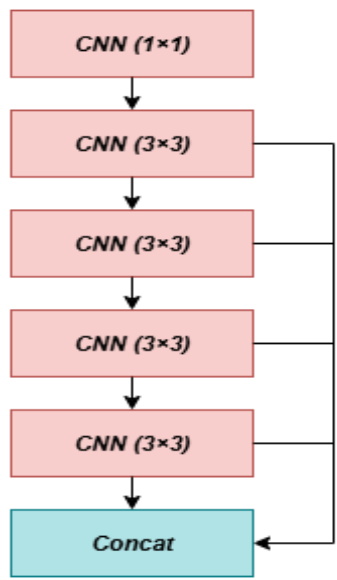

Figure 3. The schematic representation of the Bottleneck in DeepFusionNet architecture

While increasing the dimensionality of the inputs, the Upsample layer was used instead of ConvTranspose2d because the Upsample layer does not perform any convolution operation and therefore the model gained speed. The dimension is increased with Upsample layers, and as in the U-Net architecture, the corresponding encoder layer is concatenated on a channel-by-channel basis, further reducing gradient loss in the architecture. Then, using an approach similar to BasicBlocks used in the ResNet18 architecture, the CNN and activation function are used sequentially. Each Upsample layer is passed through the CBAM layer before the next step. Upsample, ResNet18 style feature extraction, as in the U-Net architecture, combining the channels from the encoder on a channel basis in the decoder and adding CBAM layers continue until the end of the decoder. The Sigmoid function at the end of the decoder compresses the values between 0 and 1. The architecture of the decoder and its connections with the CBAM layers in the encoder are shown in Figure 4. In super resolution, images are enlarged by 2x, and for this, a special layer block for super resolution is added after the last CBAM layer of the DeepFusionNet architecture. The block added to the DeepFusionNet architecture for super-resolution is shown in Figure 5.

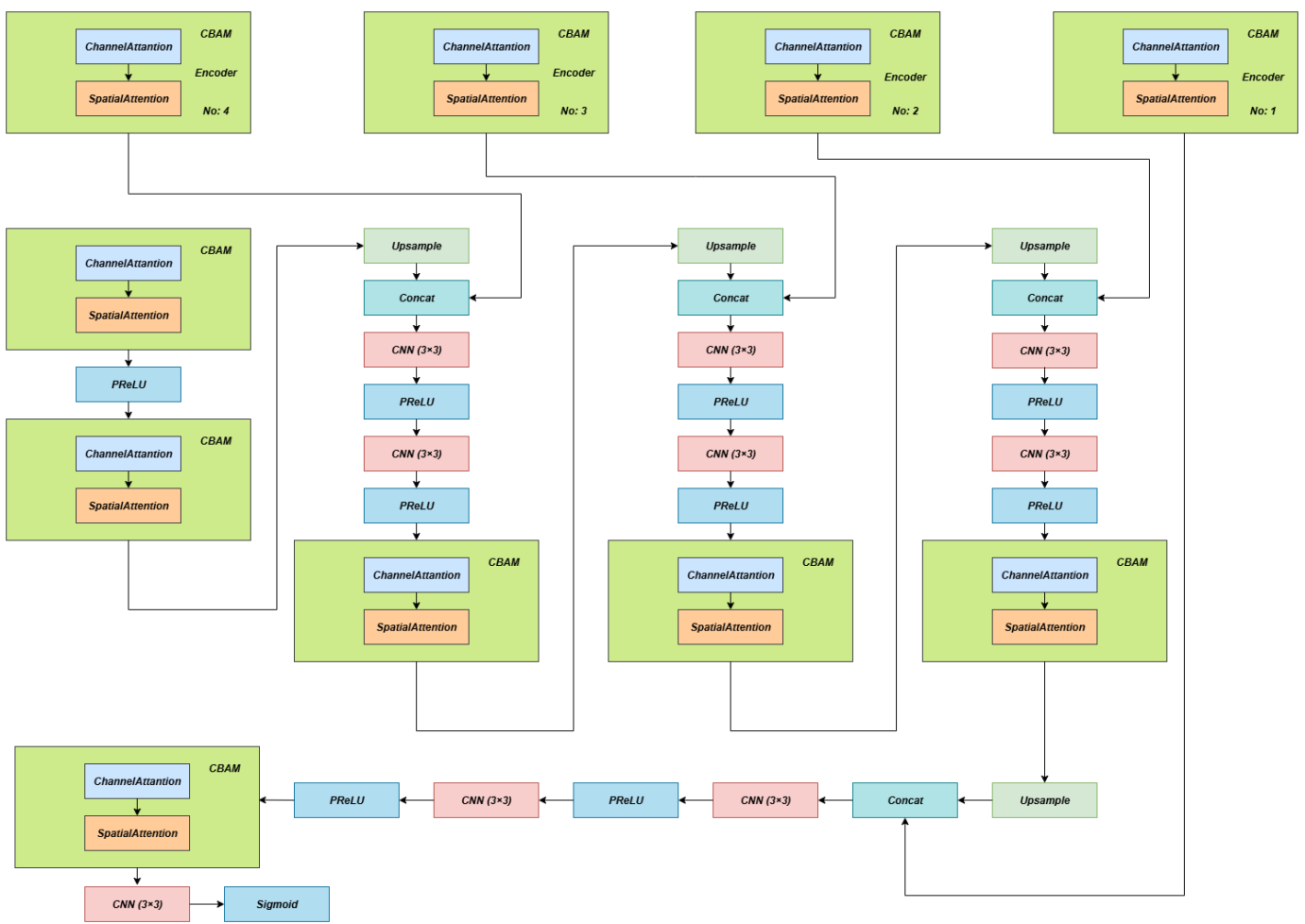

Figure 4. The architecture of the decoder and its connections with the CBAM layers in the encoder

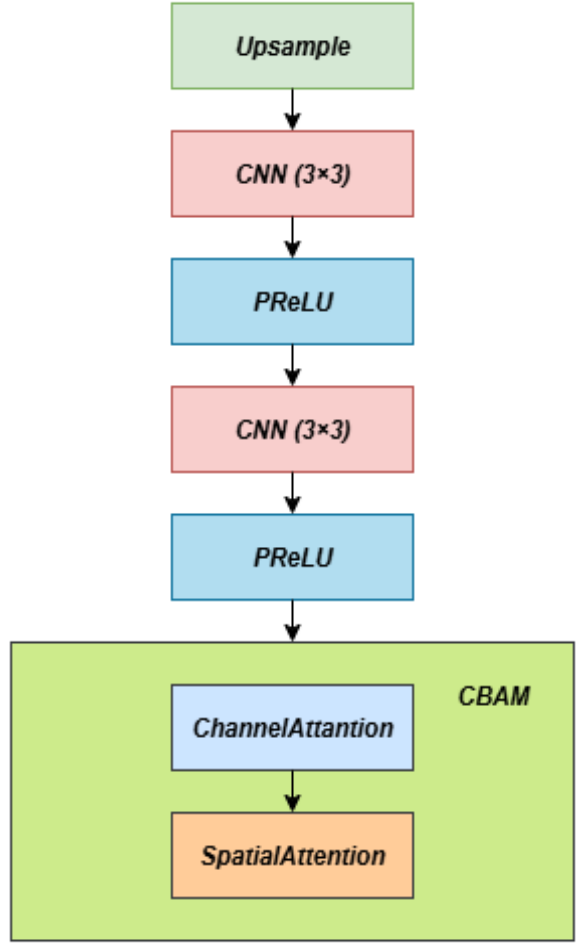

Figure 5. A block added to the DeepFusionNet architecture for super-resolution

III. RESULTS

The DeepFusionNet architecture for low-light image enhancement was trained using the LoL-v1 dataset on an A100 GPU with 40 GB of GPU RAM available at Google Colab. The model was trained for 200 epochs with a constant learning rate of 0.0001. The Adam optimizer was used instead of SGD because the Adam optimizer's momentum-driven weight updates and bias correction prevent sudden weight updates, as seen in the SGD optimizer, thus making the model training more stable [14]. The LOL-v1 dataset used during training contains 462 training and 19 validation image pairs of 256×256. The batch size parameter is set to 16, thus reducing the load on the GPU RAM. As a result of the training, it was revealed that the total number of parameters of the model was 2,597,942. This reveals that this model has even fewer parameters than YOLOv8n, demonstrating that the DeepNetFusion architecture is suitable for embedded systems. The model with DeepFusionNet architecture uses a hybrid loss function supported by MSE and SSIM. The MSE-SSIM, MSE, MAE and PSNR scores obtained by the model in the train and validation datasets over 200 epochs are shown in Figure 6. Additionally, the SSIM scores obtained by the model over 200 epochs are shown in Figure 7. For super resolution, the DeepFusionNet architecture was trained with 467 train and 24 validation image pairs. Inputs were 256x256, and blur was added using the GaussianBlur method with different kernel size values. Outputs contained blur-free images with a size of 512x512. Super-resolution model trained on A100 GPU with 0.0001 learning rate in 400 epochs. The super-resolution model contains 109,546 parameters. The low number of parameters indicates that this super-resolution model is suitable for embedded systems. The MSE-SSIM, MSE, MAE and PSNR scores obtained by the super-resolution model in the train and validation datasets over 400 epochs are shown in Figure 8.

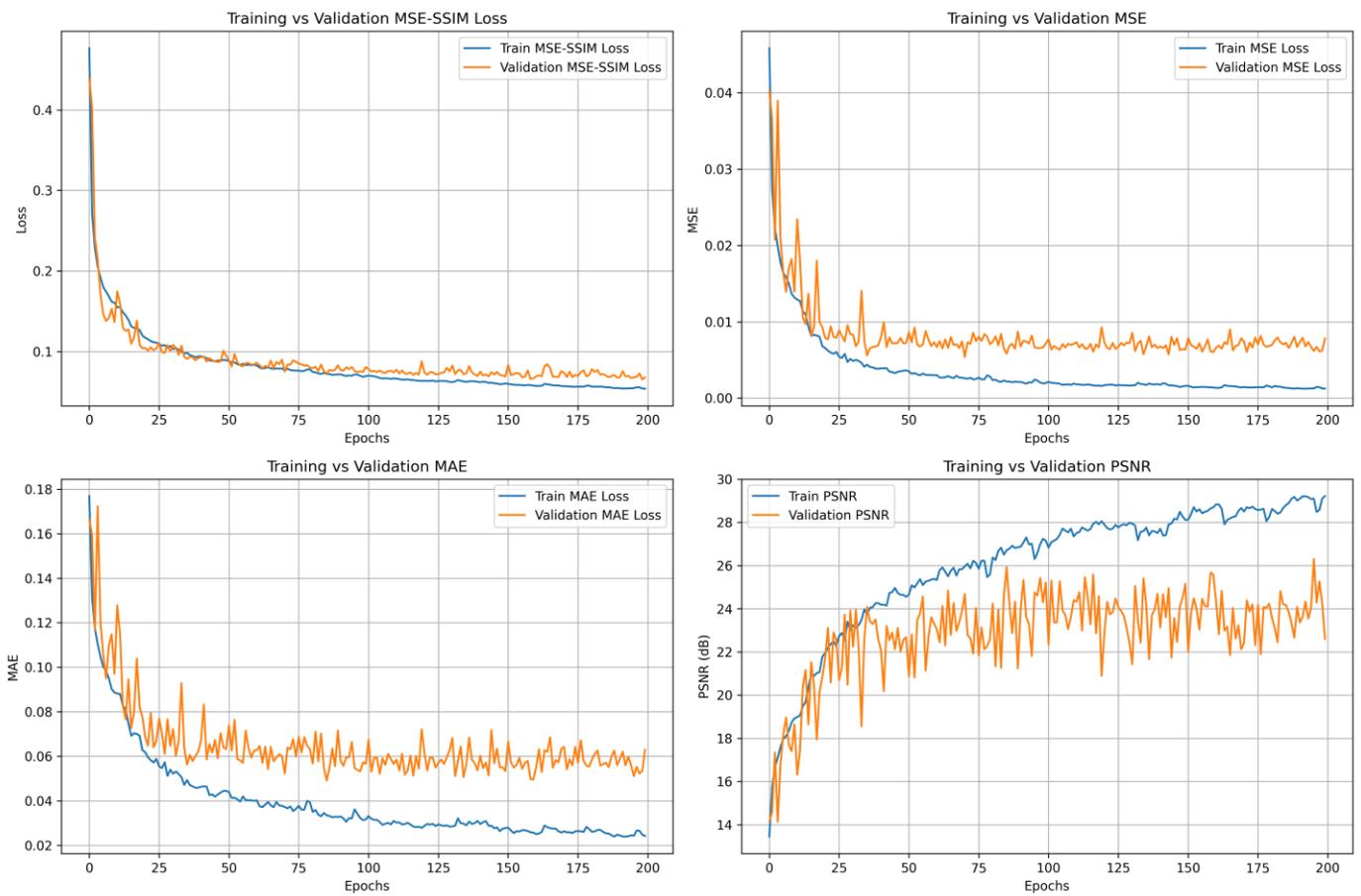

Figure 6. The graphical representation of loss functions for low-light image enhancement during epochs

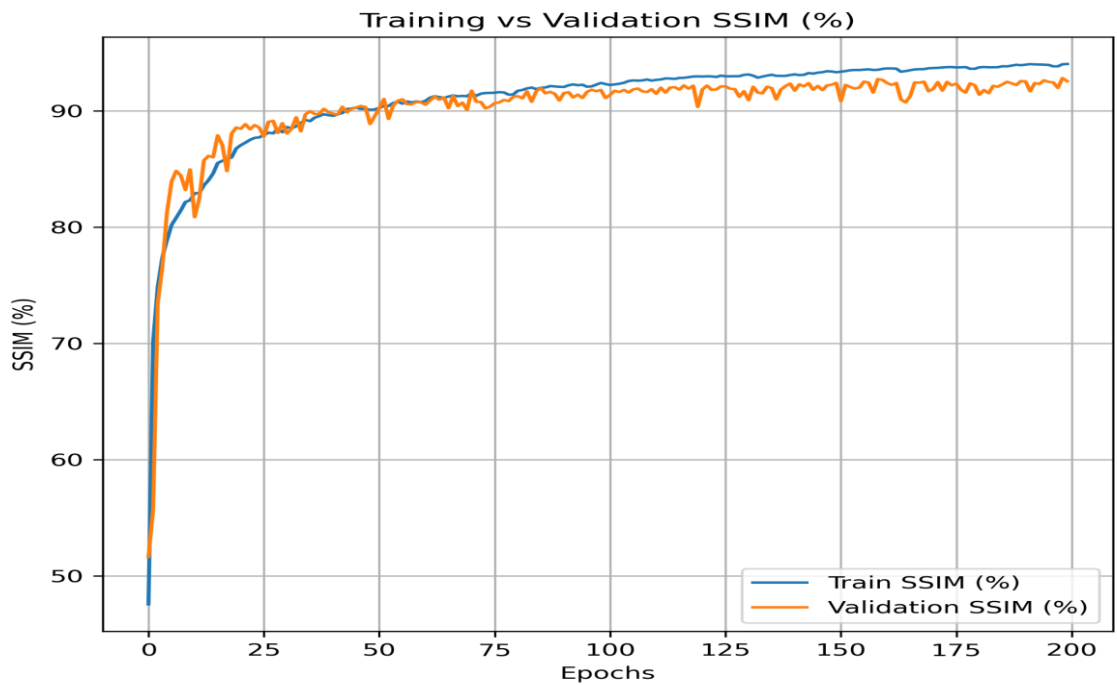

Figure 7. The graphical representation of SSIM values for low-light image enhancement during training

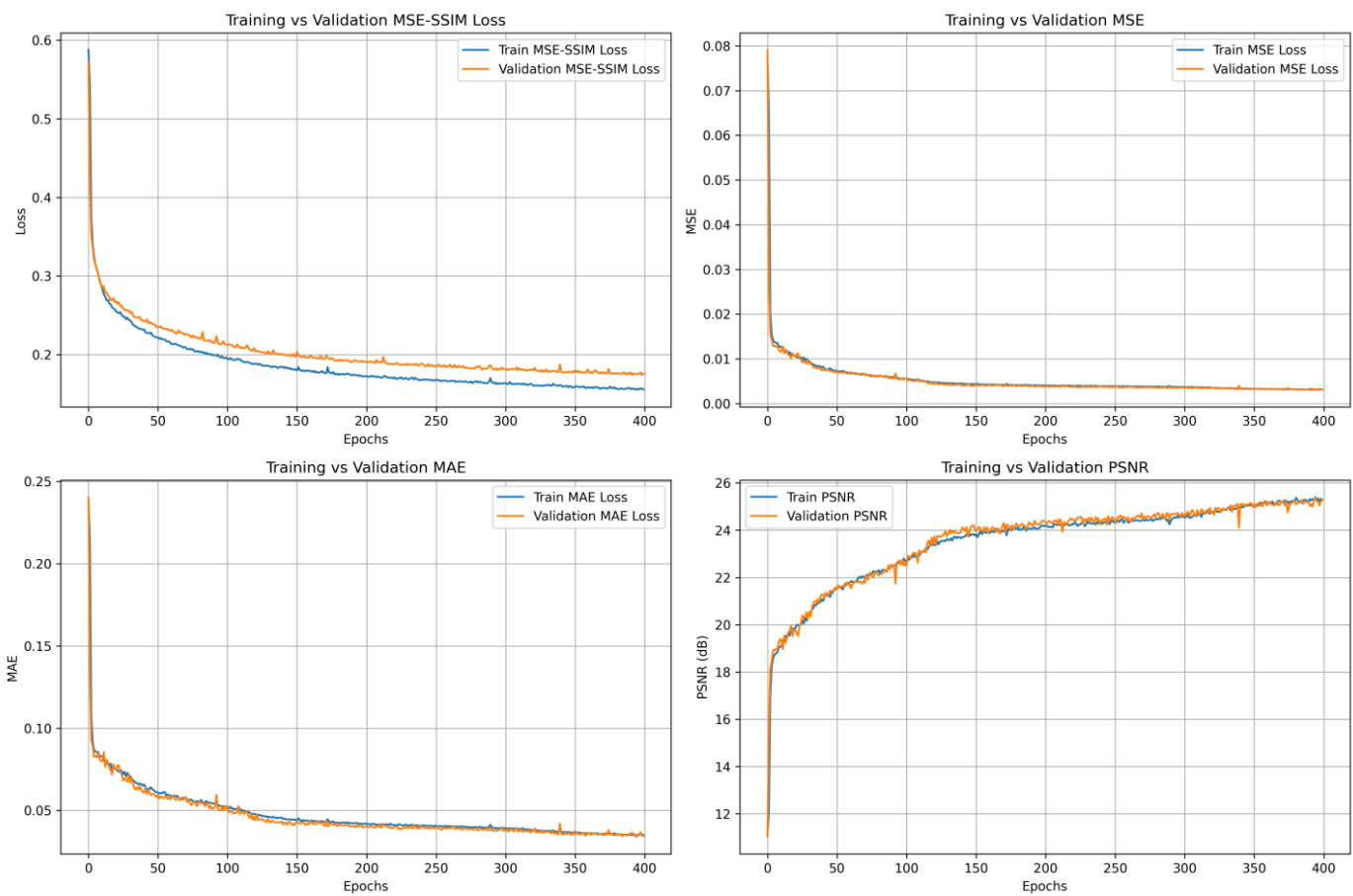

Figure 8. The graphical representation of loss functions for super-resolution during epochs

As a result of training of the DeepFusionNet architecture for low-light image enhancement, PSNR, SSIM, MSE-SSIM Loss, MSE Loss and MAE Loss results for both train and validation data are shown in Table 1. The illumination and coloring of dark images with the model created with the DeepFusionNet architecture is shown in Figure 9.

Table I. The comparison of the evaluation and loss metrics for low-light image enhancement

| Dataset Type | PSNR | SSIM (%) | MSE-SSIM Loss | MSE Loss | MAE Loss |
|---|---|---|---|---|---|
| Train | 29.2 | 92.8 | 0.05 | 0.001 | 0.02 |
| Validation | 26.3 | 94.2 | 0.06 | 0.005 | 0.04 |

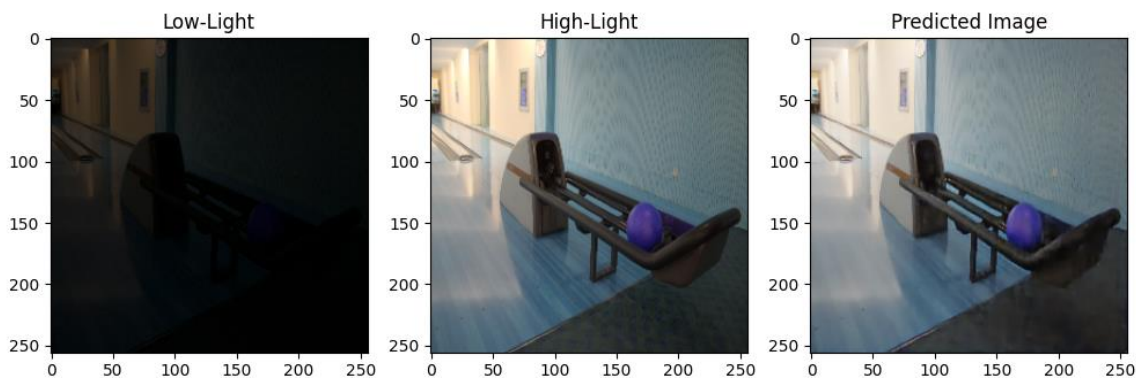

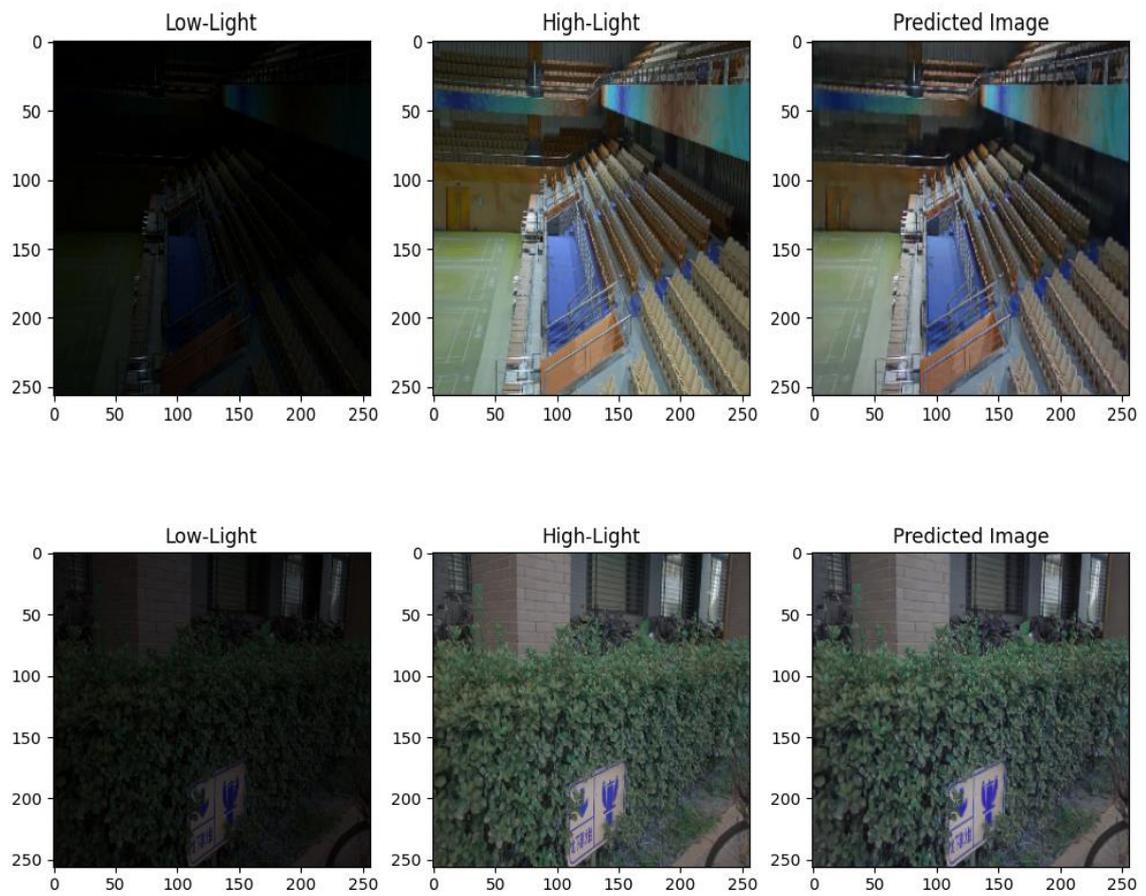

Figure 9. The illumination and coloring of dark images with the model created with the DeepFusionNet architecture

As a result of training of the DeepFusionNet architecture for super-resolution, PSNR, SSIM, MSE-SSIM Loss, MSE Loss and MAE Loss results for both train and validation data are shown in Table 2. The outputs of the super-resolution method are shown in Figure 9.

Table II. The comparison of the evaluation and loss metrics for super-resolution

| Dataset Type | PSNR | SSIM (%) | MSE-SSIM Loss | MSE Loss | MAE Loss |
|---|---|---|---|---|---|
| Train | 25.4 | 82.8 | 0.17 | 0.003 | 0.03 |
| Validation | 25.3 | 80.7 | 0.15 | 0.003 | 0.03 |

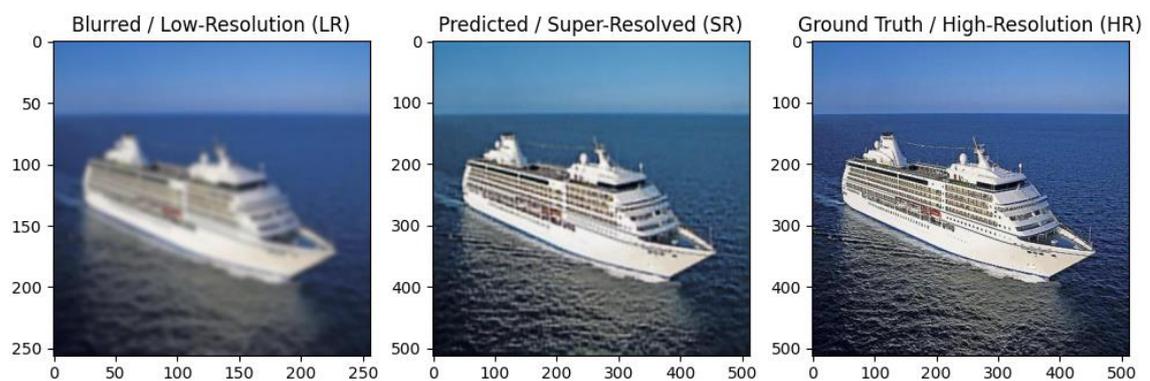

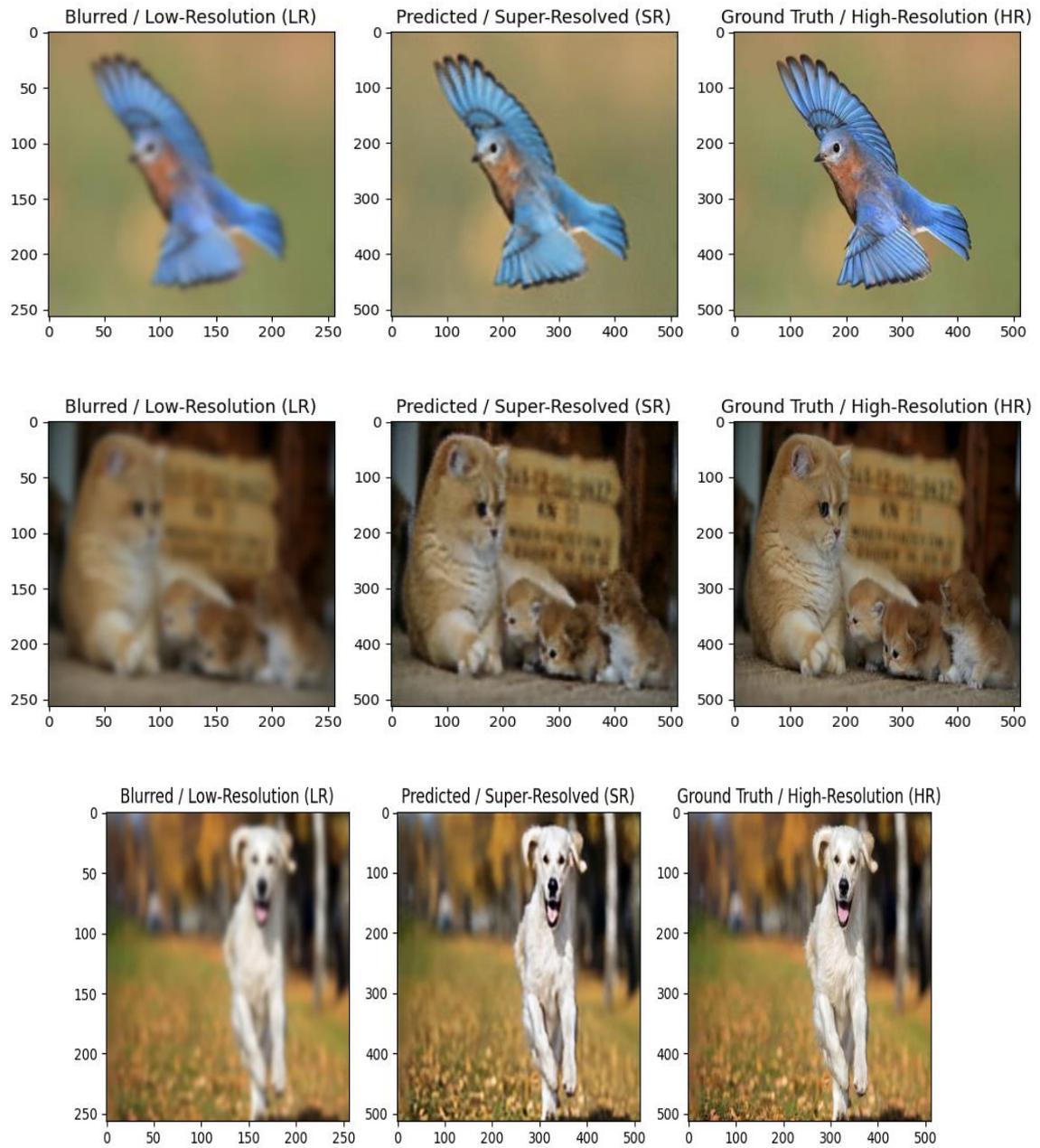

Figure 9. Super-resolution with noise reduction and 2x size increase

The comparison and visualization of SSIM scores between blurred images enhanced through super-resolution and those enlarged using standard upsampling are presented in Figure 10.

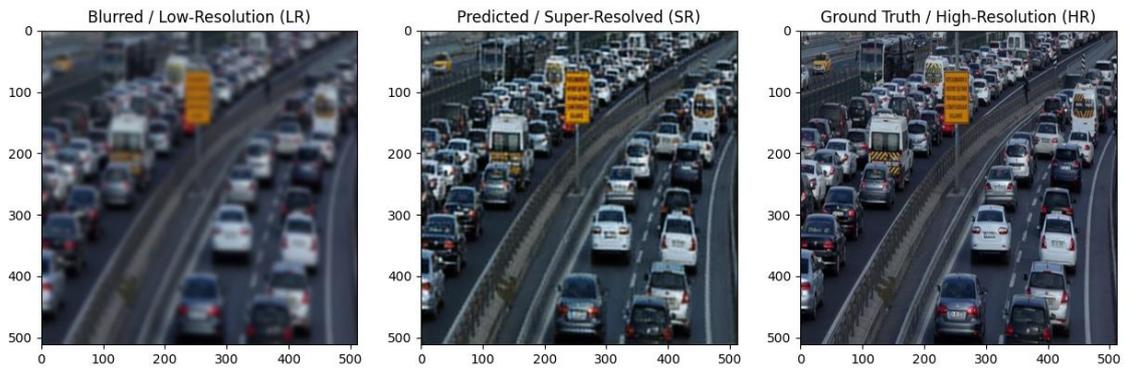

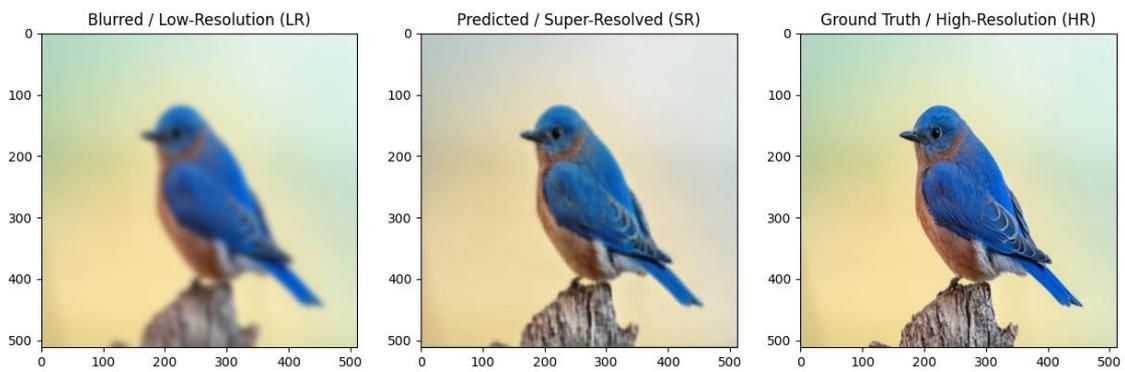

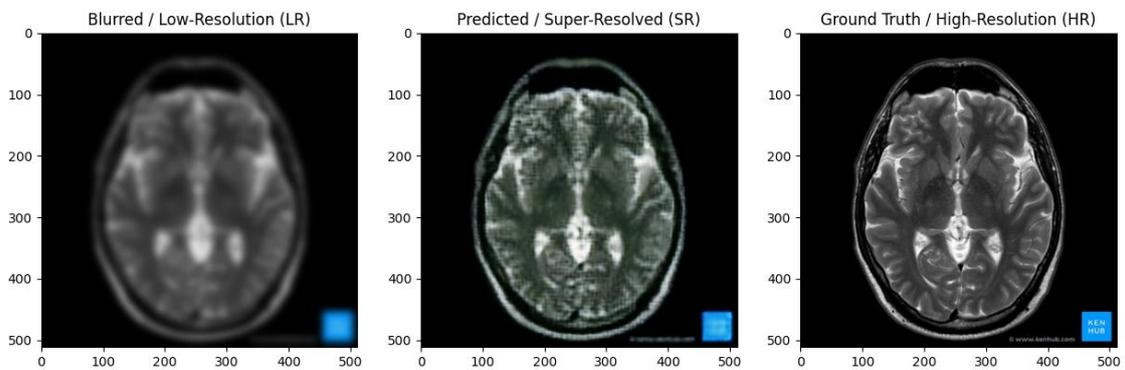

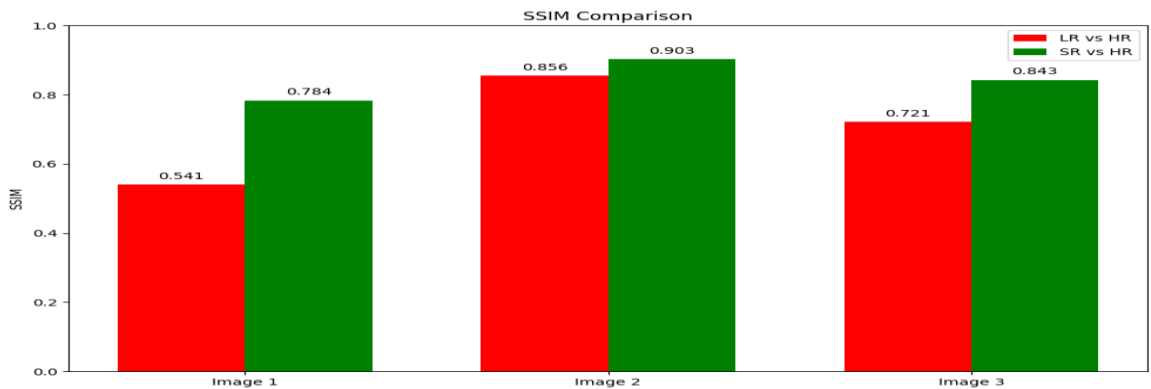

Figure 10. The comparison and visualization of SSIM scores between blurred images enhanced through super-resolution

The illumination and coloring performance and outputs of the DeepFusionNet architecture in real-time night images are shown in Figure 11 and he comparison of the DeepFusionNet model with other widely used autoencoder models is shown in Table 3.

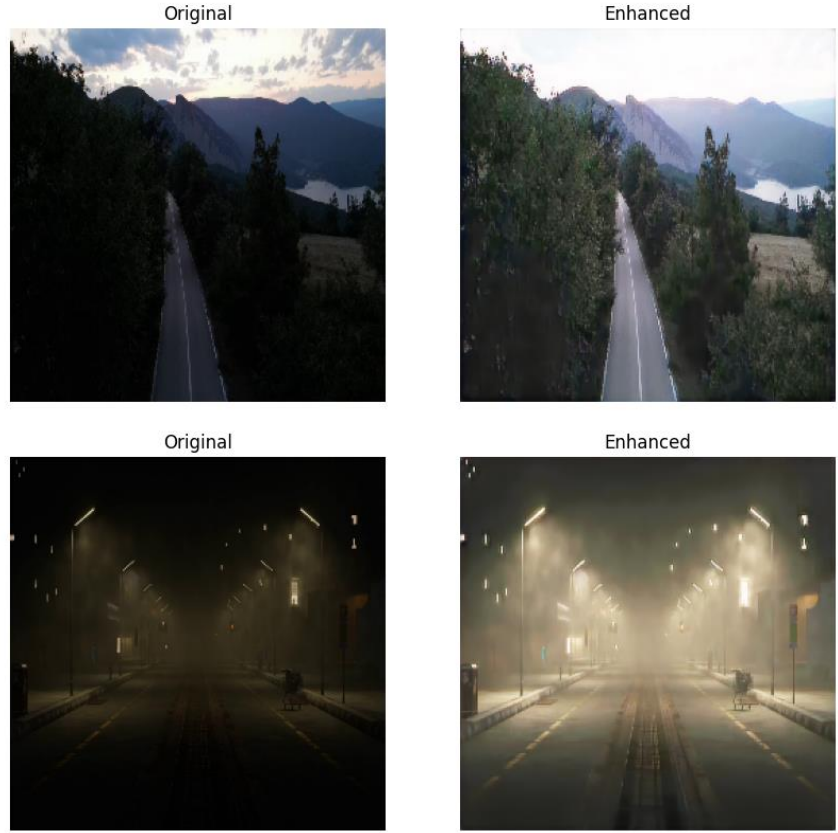

Figure 11. The illumination and coloring of real-time night and dark images with DeepFusionNet

Table III. The comparison of autoencoder models for low-light image enhancement

| Model (Architecture) | PSNR | SSIM |
|---|---|---|
| Zero-DCE++ [16] | 14.83 | 0.53 |
| GlowNet [19] | 18.29 | 0.62 |
| EnlightenGAN [17] | 19.40 | 0.82 |
| LLFlow [16] | 21.13 | 0.85 |
| LE-GAN [18] | 22.44 | 0.88 |
| RetinexNet [17] | 23.05 | 0.77 |
| ClassLIE [20] | 25.74 | 0.92 |
| DeepFusionNet (Ours) | 26.30 | 0.92 |

IV. CONCLUSIONS

As a result of this study, the DeepFusionNet architecture, which has a low number of parameters and can provide high accuracy image enhancement and super-resolution results, was introduced. The model has approximately 2.5 million parameters for low-light image enhancement, making it faster than U-Net-style architectures with approximately 30 million parameters. Furthermore, the SSIM score, which takes into account the brightness, contrast, and structural information in the images, reached a score of 92 percent in the validation dataset. On the other hand, DeepFusionNet for super-resolution achieved a PSNR score of 25.30 even though it has only 100 thousand parameters, which is approximately 32 times less parameters than YOLOv8n. These results highlight the potential of DeepFusionNet for real-time deployment in applications such as mobile photography, video surveillance, and autonomous vision systems.